\documentclass{article}
\usepackage{amsmath}
\usepackage{amsfonts}
\usepackage{amsthm}
\usepackage{amssymb}
\usepackage{bbm}
\usepackage{cuted}
\usepackage{caption}
\usepackage{hyperref}
\usepackage[capitalize]{cleveref}
\usepackage{booktabs,array,dcolumn}
\newcolumntype{d}{D{.}{.}{2.3}}
\newcolumntype{C}{>{\centering}p}
\usepackage{graphicx}
\usepackage{savesym}
\usepackage{algpseudocode}
\usepackage{float}
 \newcolumntype{L}{>{\raggedright\arraybackslash}X}
\usepackage[square,sort,comma,numbers]{natbib}
\makeatletter
\def\BState{\State\hskip-\ALG@thistlm}
\makeatother
\renewcommand{\algorithmiccomment}[1]{\hfill/\!/ #1}
\usepackage{tabularx}
\usepackage{makecell}

\newcommand{\reals}{\mathbb{R}}

\newcommand{\norm}[1]{\|#1\|}

\newcommand{\Ycal}{\mathcal{Y}}

\newcommand{\eps}{\epsilon}

\newtheorem{proposition}{Proposition}

\usepackage[accepted]{icml2018}
\usepackage{lmodern}
\usepackage{siunitx}
\usepackage{booktabs}
\usepackage{etoolbox}

\icmltitlerunning{Learning SMaLL Predictors}

\begin{document} 

\sisetup{detect-weight,mode=text}
\renewrobustcmd{\bfseries}{\fontseries{b}\selectfont}
\renewrobustcmd{\boldmath}{}
\newrobustcmd{\B}{\bfseries}

\twocolumn[
\icmltitle{Learning SMaLL Predictors}




\begin{icmlauthorlist}
\icmlauthor{Vikas K. Garg}{to}
\icmlauthor{Ofer Dekel}{goo}
\icmlauthor{Lin Xiao}{goo}
\end{icmlauthorlist}

\icmlaffiliation{to}{CSAIL, MIT, Cambridge, MA, USA.}
\icmlaffiliation{goo}{Microsoft Research, Redmond, WA, USA}

\icmlcorrespondingauthor{Vikas K. Garg}{vgarg@csail.mit.edu}

\icmlkeywords{boring formatting information, machine learning, ICML}

\vskip 0.3in
]



\printAffiliationsAndNotice{}  

\begin{abstract} 
We present a new machine learning technique for training small resource-constrained predictors. Our algorithm, the Sparse Multiprototype Linear Learner (SMaLL), is inspired by the classic machine learning problem of learning $k$-DNF Boolean formulae. 
We present a formal derivation of our algorithm and demonstrate the benefits of our approach with a detailed empirical study. 
\end{abstract} 

\section{Introduction}

Modern advances in machine learning have produced models that achieve unprecedented accuracy on standard benchmark prediction problems. However, this remarkable progress in accuracy has come at a significant computational cost. Many state-of-the-art machine learned models have ballooned in size and running one of these models on a new data point can require tens of GFLOPs. On the other hand, there is also a renewed interest in developing machine learning techniques that produce small models, which can run effectively on resource-constrained platforms, like smart phones, wearables, and other devices \cite{Gupta17}. Instead of pursuing predictive accuracy at all costs, the goal is to investigate and understand the cost-accuracy tradeoff. In many application domains, an engineer designing an embedded AI system will gladly accept a small decrease in accuracy in exchange for a drastic reduction in computational cost.

In addition to the high computational cost, the recent progress in accuracy has also come at the expense of model interpretability. Massive deep neural networks, large ensembles of decision trees, and other state-of-the-art models are black box predictors, which do not offer any intuitive explanation for their predictions. Black box predictors can be dangerous, as they may hide unintended biases and behave in unexpected ways. The size of a model and its interpretability are often negatively correlated, simply because humans are not good at reasoning about large and complex objects.

In pursuit of compact and interpretable machine-learnable models, we take inspiration from the classic machine learning paradigm of learning Boolean formulae. Specifically, we revisit the old idea of learning \emph{disjunctive normal forms} (DNFs). A DNF, also known as a \emph{disjunction of conjunctions}, is a logical disjunction of one or more logical conjunctions of one or more variables. In other words, a DNF is an \emph{or} of \emph{and}'s. For example, the formula 
$$
(A \land \lnot B \land C) \lor (\lnot A \land B \land C \land D) \lor (\lnot C \land \lnot D)~~,
$$
is a DNF over the Boolean variables $A,B,C,D$. More specifically, a $p$-term $k$-DNF is a DNF with $p$ terms, where each term contains exactly $k$ variables. Learning $k$-DNFs is one of the classic problems in learning theory, originally discussed in Valiant's seminal paper on the PAC learning model \cite{Valiant84}. While a handful of previous papers propose heuristic practical approaches to learning $k$-DNFs \cite{Cord01,Hauser10,Wang15}, most of the previous work on this problem is of pure theoretical interest. One rich line of research focuses on theoretically characterizing the difficulty of learning a $k$-DNF in various unrealistic and restricted models of learning \cite{Verbeurgt90,Blum94,Mansour95,Blum95,Jackson97,Verbeurgt98,Bshouty99,Sakai00,Servedio04,Bshouty05,Feldman12}. Another line of theoretical work addresses the (unrestricted) problem of PAC learning a $k$-DNF, but basically shows that this problem is extremely difficult \cite{Bshouty96,Tarui99,Klivans04,Khot08}. We emphasize that our paper does not make any progress on the fundamental theory of learning $k$-DNFs, and instead uses $k$-DNFs to inspire practical algorithms that learn small and interpretable models.

Small DNFs are a natural starting point for our research, because they pack a powerful nonlinear descriptive capacity in a very small form factor. The DNF structure is also known to be intuitive and interpretable by humans \cite{Hauser10,Wang15}. For example, imagine that a bank uses a DNF formula to determine whether or not to approve a mortgage application. Specifically, say that the model approves an application if the applicant has a credit score above 630 and makes a 25\% downpayment, OR if she has resided at her current address for at least three years, is steadily employed, and earns more than 60K a year. This DNF rule is easy to understand and explain, as long as $k$ (the number of variables in each term) is small. 

While the merits of $k$-DNFs have been known for a long time, the problem has always been that they are incredibly difficult to learn from labeled training data. We bypass this problem by focusing on a continuous relaxation of $k$-DNFs, which we call \emph{Sparse Multiprototype Linear Predictors}. Start with a $p$-term $k$-DNF defined over a set of $n$ Boolean variables. Encode the $j$'th term in the DNF formula by a vector $w_j \in \{-1,0,1\}^n$, where 
\begin{equation}
    w_{j,l} = 
    \begin{cases} 
        1 & \text{$l$'th variable appears as positive} \\
        -1 & \text{$l$'th variable appears as negative} \\
        0 &\text{$l$'th variable doesn't appear} 
    \end{cases}~~.
\end{equation}
Notice that the resulting vector is $k$-sparse. Next, let $x \in \{-1,1\}^n$ encode the Boolean assignment of the input variables, where $x_l = 1$ encodes that the $l$'th variable is true and $x_l = -1$ encodes that it is false. Note that the $j$'th term of the DNF is satisfied if and only if $w_j \cdot x \geq k$. Moreover, note that the entire DNF is satisfied if and only if 
\begin{equation}\label{eqn:kdnfRelax}
\max_{j \in [p]}~  w_j \cdot x  \geq k~~,
\end{equation}
where we use $[p]$ as shorthand for the set $\{1,\ldots,p\}$. We relax this definition by allowing the input $x$ to be an arbitrary vector in $\reals^n$ and allowing each $w_j$ to be any $k$-sparse vector in $\reals^n$. By construction, the class of models of this form is at least as powerful as the original class of $p$-term $k$-DNF Boolean formulae. Therefore, learning this class of models is a form of improper learning of $k$-DNFs.

After allowing $x$ and $w_j$ to take arbitrary real values, the threshold $k$ on the right-hand side of \eqref{eqn:kdnfRelax} becomes somewhat arbitrary, so we replace it with zero. Overall, a $k$-sparse $p$-prototype linear predictor is a function of the form
$$
f(x) ~=~ \text{sign} \left( \max_{j\in[p]}~ w_j \cdot x \right)~~.
$$
It is interesting to note that other recent advances in resource constrained prediction were also achieved by revisiting classic machine learning paradigms. Specifically, the resource constrained prediction technique presented in \cite{Gupta17} is a modern take on nearest neighbor classifiers, and the technique presented in \cite{Kumar17} is an sophisticated enhancement of a small decision tree. 

In Section \ref{sec:setup}, we formulate the problem of learning a $k$-sparse $p$-prototype linear predictor as a mixed integer nonlinear optimization problem. Then, in Section \ref{sec:saddle-point}, we relax this optimization problem to a saddle-point problem, which we solve using a Mirror-Prox algorithm. We name the resulting algorithm \emph{Sparse Multiprototype Linear Learner}, or SMaLL for short. Finally, we present empirical results that demonstrate the merits of our approach in Section \ref{sec:experiments}. All proofs are provided in the Supplementary to improve readability. 

\section{Problem Formulation} \label{sec:setup}


We first derive a convex loss function for multi-prototype
binary classification. Let $\{(x_i,y_i)\}_{i=1}^m$ be a training set of instance-label pairs, where each $x_i \in \reals^n$ and each $y_i \in \{-1,1\}$. Let $\ell:\reals \mapsto \reals$ be a convex surrogate for the 
error indicator function 
$\mathbbm{1}_{f(x_i)\neq y_i}=1$ if $y_i f(x_i)<0$ and $0$ otherwise.
Other than being convex, we also assume that~$\ell$ upper bounds the 
error indicator function and is monotonically non-increasing.
In particular, the popular hinge-loss and log-loss functions all
satisfy these properties. 

We handle the positive and the negative examples separately. 
For each negative training example $(x_i,-1)$, the classifier makes a correct
prediction if and only if $\max_{j \in [p]} w_j\cdot x_i < 0$. 
Under our assumptions on~$\ell$, 
the error indicator function can be upper bounded as
$$
\mathbbm{1}_{f(x_i) \neq -1}
\leq
\ell\Bigl( -\max_{j \in [j]} w_j\cdot x_i  \Bigr) 
=\max_{j \in [p]} \ell( - w_j\cdot x_i ),
$$ 
where the equality holds because we assume 
that~$\ell$ is monotonically non-increasing. 
We note that the upper bound $\max_{j \in [p]} \ell( - w_j\cdot x_i )$
is jointly convex in $\{ w_j \}_{j=1}^p$
\citep[Section~3.2.3]{BoydVandenberghe2004book}.

For each positive example $(x_i,+1)$, the classifier makes a
correct prediction if and only if $\max_{j \in [p]} w_j\cdot x_i > 0$.
By our assumptions on~$\ell$, we have
\begin{equation}\label{eqn:positive-ub}
\mathbbm{1}_{f(x_i) \neq +1}
\leq
\ell\Bigl( \max_{j \in [p]} w_j\cdot x_i  \Bigr) 
=\min_{j\in[p]} \ell(w_j\cdot x_i).
\end{equation}
Again, the equality above is due to the monotonic non-increasing property
of~$\ell$.
Here the right-hand side $\min_{j\in[p]} \ell(w_j\cdot x_i)$
is not convex in the prototype vectors $\{w_j\}_{j=1}^p$.
We resolve this by designating a dedicated prototype $w_{j(i)}$
for each positive training example $(x_i,+1)$, and use the looser upper bound 
$$
\mathbbm{1}_{f(x_i) \neq +1}
\leq \ell(w_{j(i)}\cdot x_i).
$$ 
In the extreme case, we can associate each positive example with a distinct
prototype, then there will be no loss of using $\ell(w_{j(i)}\cdot x_i)$
compared with the upper bound in~\eqref{eqn:positive-ub}, 
by setting $j(i)=\arg\max_{j\in[p]} w_j\cdot x_i$.
However, in this case, the number of prototypes~$p$ is equal to the number of 
positive examples, which can be excessively large for storage and computation
as well as cause overfitting.
In practice, we can cluster the positive examples into~$p$ groups, 
where~$p$ is much smaller than the number of positive examples, 
and assign all positive examples in each group with a common prototype. 
In other words, 
we have $j(i)=j(k)$ if $x_i$ and~$x_k$ belong to the same cluster.

Overall, we have the following convex surrogate for the total number of 
training errors:
\begin{align}
h(w_1,\ldots,w_s)
=& \sum_{i\in I_+} \ell\big(w_{j(i)}\cdot x_i  \big) \nonumber \\
&+ \sum_{i\in I_-} \max_{j\in[p]} \ell( - w_j\cdot x_i),
\label{eqn:overall-loss}
\end{align}
where $I_+=\{i:y_i=+1\}$ and $I_-=\{i:y_i=-1\}$.
In the rest of this paper, we let $W \in \mathbb{R}^{p \times n}$ be the matrix 
formed by stacking the vectors $w_1^T, \ldots, w_p^T$ vertically,
and denote the above loss function by $h(W)$.
In order to train a multi-prototype classifier, we minimize the regularized
surrogate loss:
\begin{equation}\label{eqn:convex-regu-loss}
\min_{W\in\reals^{p\times n}}~ \frac{1}{m} h(W) + \frac{\lambda}{2}\|W\|_F^2,
\end{equation}
where $\|\cdot\|_F$ denotes the Frobenius norm of a matrix.

In this paper, we focus on the log-loss 
$$\ell(z)=\log(1+\exp(-z)).$$
Although this~$\ell$ is a smooth function, the overall loss~$h$
defined in~\eqref{eqn:overall-loss} is non-smooth, due to the 
$\max$ operator in the sum over $I_-$.
In order to take advantage of fast algorithms for smooth optimization,
we smooth the loss function using soft-max. 

\subsection{Smoothing the Loss via Soft-Max}
More specifically, we replace the non-smooth terms 
$\max_{j\in[p]}\ell(t_j)$ with
\begin{equation}\label{eqn:soft-max}
u(t) \triangleq \log\biggl(1+\sum_{j\in[p]} \exp(-t_j)\biggr)~,
\end{equation}
where $t=(t_1,\ldots,t_p)\in\reals^p$.
Then we obtain the smoothed loss function
\begin{equation}\label{eqn:smoothed-loss}
\tilde{h}(W) 
= \sum_{i\in I_+} \ell\big(w_{j(i)}\cdot x_i  \big) 
+ \sum_{i\in I_-} u( W x_i),
\end{equation}
around which we will customize our algorithm design. We now incorporate sparsity constraints explicitly for the prototypes
$w_1, \ldots, w_p$. 

\subsection{Incorporating Sparsity via Binary Variables}

With some abuse of notation, we let $\|w_j\|_0$ denote the number of non-zero
entries of the vector $w_j$, and define
\[
  \|W\|_{0,\infty} \triangleq \max_{j\in[p]} \|w_j\|_0 .
\]
The requirement that each prototype is $k$-sparse translates into the constraint
$\|W\|_{0,\infty}\leq k$.
Therefore the problem of training a SMaLL model with
budget~$k$ (for each prototype) can be formulated as
\begin{equation}\label{eqn:K-DNF-primal}
\min_{\substack{{W \in \mathbb{R}^{p \times n}}\\ ||W||_{0, \infty} \leq k}} 
\frac{1}{m}\tilde{h}(W) +  \dfrac{\lambda}{2} ||W||_F^2 ~,
\end{equation}
where $\tilde{h}$ is defined in~\eqref{eqn:smoothed-loss}.
This is a very hard optimization problem due to the nonconvex sparsity 
constraint.

Following the approach of \citet{PilanciWainwright2015Boolean},
we introduce a binary matrix $\epsilon \in \{0, 1\}^{p \times n}$
and rewrite \eqref{eqn:K-DNF-primal} as
\begin{equation*}
\min_{\substack{{W \in \mathbb{R}^{p \times n}}\\ \epsilon \in \{0, 1\}^{p \times n} \\ \|\epsilon\|_{1, \infty} \leq k}} 
\dfrac{1}{m} \tilde{h}(W \odot \epsilon) + \dfrac{\lambda}{2} ||W \odot \epsilon||_F^2~,
\end{equation*}
where $\odot$ denotes the Hadamard product (entry-wise products) of two matrices.
Here 
$$\|\epsilon\|_{1,\infty}=\max_{j\in[p]}\|\epsilon_j\|_1~,$$ 
where $\epsilon_j$ is the $j$th row of~$\epsilon$.
Since all entries of~$\epsilon$ belong to $\{0,1\}$, 
the constraint $\|\epsilon\|_{1,\infty}\leq k$ is the same as
$\|\epsilon\|_{0,\infty}\leq k$.
Noting that we can take $W_{ij}=0$ when $\epsilon_{ij}=0$ and vice-versa, 
this problem is equivalent to
\begin{equation}\label{eqn:boolean-primal}
\min_{\substack{{W \in \mathbb{R}^{p \times n}}\\ \epsilon \in \{0, 1\}^{p \times n} \\  ||\epsilon||_{1, \infty} \leq k}} 
\dfrac{1}{m} \tilde{h}(W \odot \epsilon) + \dfrac{\lambda}{2} ||W||_F^2 ~.
\end{equation}
Using~\eqref{eqn:smoothed-loss}, the objective function can be written as
\[
\frac{1}{m}\biggl(\sum_{i\in I_+}\!\ell\bigl((W\odot\epsilon)_{j(i)} x_i\bigr) 
+ \sum_{i\in I_-}\!\!u\bigl(-(W\odot\epsilon) x_i\bigr)\!\biggr) 
+\frac{\lambda}{2}\|W\|_F^2,
\]
where $(W\odot\epsilon)_{j(i)}$ denotes the $j(i)$th row of $W\odot\epsilon$.

\begin{figure*}
\centering
\includegraphics[width=0.95\textwidth]{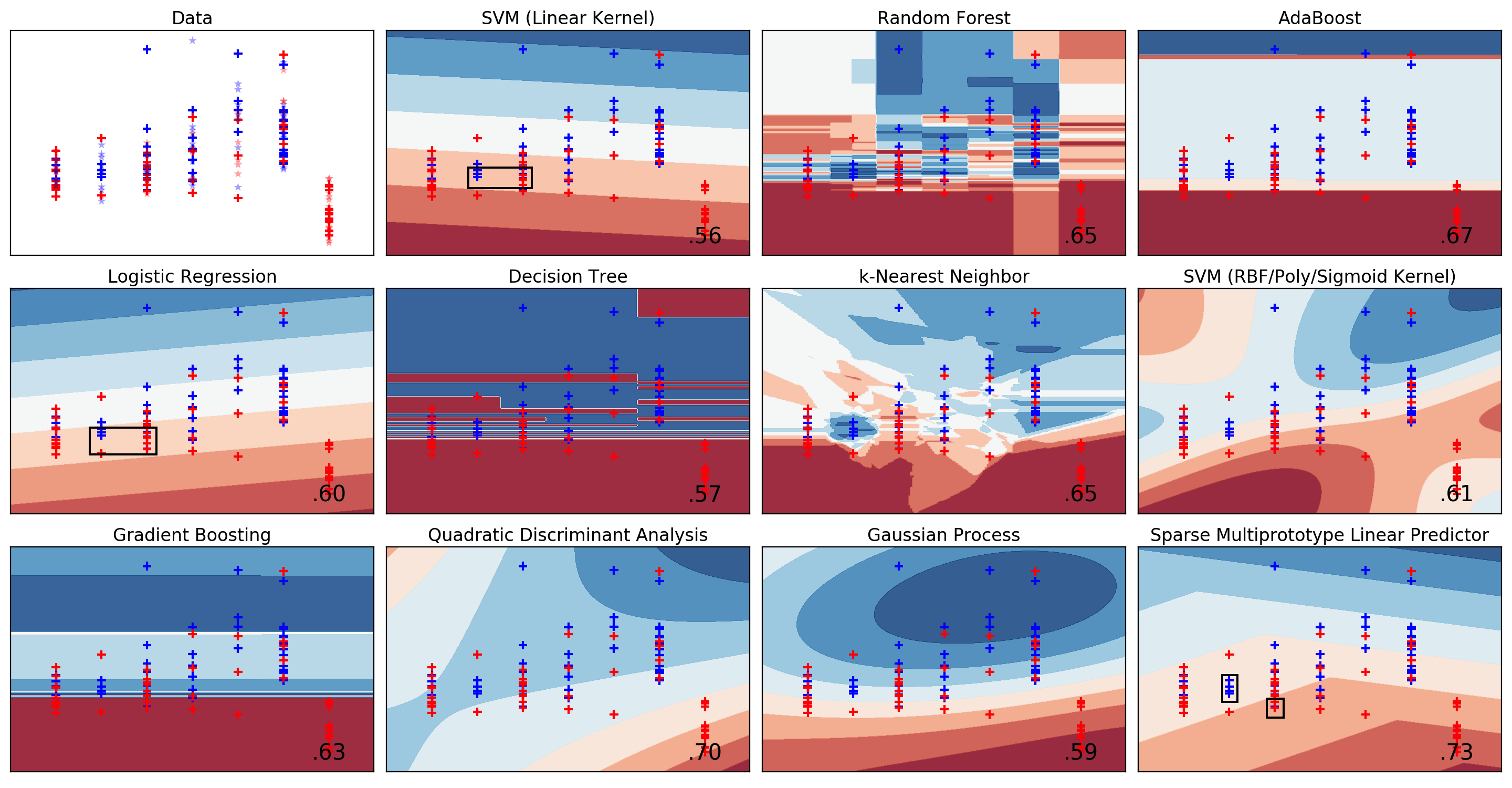}
\caption{Decision surfaces of different classifier types on the two-dimensional \emph{chscase funds} toy dataset. Test classification accuracy is shown on the bottom right of each plot.}
\label{fig1:sleuth}
\end{figure*}

We now derive a saddle-point formulation of the mixed-integer
nonlinear optimization problem~\eqref{eqn:boolean-primal}. We can then derive
a convex-concave relaxation that can be solved efficiently by
the Mirror-Prox algorithm.

\section{Saddle-Point Relaxation}\label{sec:saddle-point}
We show that problem in~\eqref{eqn:boolean-primal}
is equivalent to the following minimax saddle-point problem:
\begin{equation}\label{eqn:min-max-Phi}
\min_{\substack{{W \in \mathbb{R}^{p \times n}}\\ \epsilon \in \{0, 1\}^{p \times n} \\  ||\epsilon||_{1, \infty} \leq k}} 
~ \max_{\substack{S=[s_1\cdots s_m] \\ s_i\in\mathcal{S}_i, ~i\in[m]}}
~ \Phi(W, \epsilon, S),
\end{equation}
where $S\in\reals^{p\times m}$,
each of its column $s_i$ belongs to a set $\mathcal{S}_i\subset\reals^p$ 
(which are given in Proposition~\ref{prop:conjugate}),
and
\[
\Phi(W, \epsilon, S) \!=\! \frac{1}{m} \!\! 
\sum_{i\in[m]} \!\Bigl(y_i s_i^T(W\odot\epsilon)x_i - u^\star(s_i)\!\Bigr)
 + \dfrac{\lambda}{2} ||W||_F^2 .
\]
Here $u^\star$ is the convex conjugate of $u$ defined in~\eqref{eqn:soft-max}:
\begin{align}
&u^\star(s_i)=\sup_{t\in\reals^p}\bigl\{s_i^T t-u(t)\bigr\} 
\label{eqn:u-conjugate}\\
&= \begin{cases} \displaystyle 
\sum_{j=1}^p (-s_{i,j})\log(-s_{i,j}) + (1\!+\!{\bf 1}^T s_i) \log (1\!+\!{\bf 1}^T s_i), \\
\qquad\quad \mbox{if } s_{i,j} \leq 0 ~\forall j \in [p]$ and ${\bf 1}^T s_i \geq -1; \\
\infty, \hspace*{0.4cm} \mbox{ otherwise }.  
\end{cases}
\nonumber
\end{align}

The equivalence between~\eqref{eqn:boolean-primal} and~\eqref{eqn:min-max-Phi}
is a direct consequence of the following proposition. 

\begin{proposition} \label{prop:conjugate}
Let $\ell(z) = \log(1 + \exp(-z))$ where $z\in\reals$
and $u(t)=\log\bigl(1+\sum_{j\in[p]}\exp(-t_j)\bigr)$ where $t\in\reals^p$. 
Then for $i\in I_+$, we have
\begin{align*}
\ell\bigl((W\odot\epsilon)_{j(i)} x_i\bigr) 
&= \max_{s_i\in \mathcal{S}_i} \left(y_i s_i^T(W\odot\epsilon) x_i - u^\star(s_i) \right), 
\end{align*}
where $y_i=+1$ and 
\[
\mathcal{S}_i
=\Bigl\{s_i\in\reals^p : s_{i,j(i)}\!\in\![-1,0],~s_{i,j}\!=\!0 ~\forall j\neq\! j(i)\Bigr\}.
\]
For $i\in I_-$, we have 
\begin{align*}
u\bigl(-(W\odot\epsilon) x_i\bigr) 
&= \max_{s_i\in \mathcal{S}_i} \left(y_i s_i^T(W\odot\epsilon) x_i - u^\star(s_i) \right) ,
\end{align*}
where $y_i=-1$ and
\[
\mathcal{S}_i 
=\Bigl\{s_i\in\reals^p : {\bf 1}^T s \geq -1, ~s_{i,j}\leq 0 ~\forall j\in[p]\Bigr\}.
\]
\end{proposition}

We can further eliminate the variable~$W$ in~\eqref{eqn:min-max-Phi}, which
is facilitated by the following result.

\begin{proposition}\label{prop:eliminate-W}
For any given $\epsilon\in\{0,1\}^{p\times n}$ and 
$S\in\mathcal{S}_1\times\cdots\times\mathcal{S}_m$, the solution to
\[
\min_{W\in\reals^{p\times n}} \Phi(W, \epsilon, S)
\]
is unique and given by 
\begin{equation}\label{eqn:optimal-W}
W(\epsilon, S) 
= -\frac{1}{m\lambda}\sum_{i\in[m]} y_i \bigl(s_i x_i^T\bigr)\odot\epsilon. 
\end{equation}
\end{proposition}
Now we substitute $W(\epsilon, S)$ into~\eqref{eqn:min-max-Phi} to obtain
\begin{equation}\label{eqn:min-max-phi}
\min_{\substack{\epsilon \in \{0, 1\}^{p \times n} \\ \|\epsilon\|_{1, \infty} \leq k}} 
~ \max_{\substack{S=[s_1\cdots s_m] \\ s_i\in\mathcal{S}_i, ~i\in[m]}}
~ \phi(\epsilon, S),
\end{equation}
where 
\[
\phi(\epsilon, S) = -\frac{1}{m\lambda} 
\biggl\|\sum_{i\in[m]}y_i\bigl(s_i x_i^T\bigr)\odot\epsilon\biggr\|_F^2 
-\sum_{i\in[m]} u^\star(s_i).
\]
The above expression of $\phi(\epsilon,S)$ is concave in~$S$ 
(which is to be maximized), but not convex in~$\epsilon$ 
(which is to be minimized).
However, because $\epsilon\in\{0,1\}^{p\times n}$, 
\begin{align*}
&\biggl\|\sum_{i\in[m]}y_i\bigl(s_i x_i^T\bigr)\odot\epsilon\biggr\|_F^2\\ 
=& \sum_{i\in[m]}y_i s_i^T \biggl(
\sum_{i\in[m]}y_i\bigl(s_i x_i^T\bigr)\odot\epsilon\odot\epsilon\biggr)x_i\\
=& \sum_{i\in[m]}y_i s_i^T \biggl(
\sum_{i\in[m]}y_i\bigl(s_i x_i^T\bigr)\odot\epsilon\biggr)x_i,
\end{align*}
where we used $\epsilon\odot\epsilon=\epsilon$. Therefore we have
\begin{align}
\phi(\epsilon, S) 
=& -\frac{1}{m\lambda} \sum_{i\in[m]}y_i s_i^T \biggl(
\sum_{i\in[m]}y_i\bigl(s_i x_i^T\bigr)\odot\epsilon\biggr)x_i \nonumber\\
&-\sum_{i\in[m]} u^\star(s_i),
\label{eqn:convex-concave-phi}
\end{align}
which is concave in~$S$ and linear (thus convex) in~$\epsilon$.

Finally, we relax the integrality constraint on~$\epsilon$
to its convex hull, i.e., $\epsilon\in[0,1]^{p\times n}$, and consider 
\begin{equation}\label{eqn:min-max-phi-relax}
  \min_{\substack{\epsilon \in [0, 1]^{p \times n} \\ \|\epsilon\|_{1, \infty} \leq k}} 
~ \max_{\substack{S=[s_1\cdots s_m] \\ s_i\in\mathcal{S}_i, ~i\in[m]}}
~ \phi(\epsilon, S),
\end{equation}
where $\phi(\epsilon,S)$ is given in~\eqref{eqn:convex-concave-phi}.
This is a convex-concave saddle-point problem, which can be solved
efficiently (in polynomial time) by the Mirror-Prox algorithm
\citep{Nemirovski2004mirror-prox,JuditskyNemirovski2011chapter6}.

After finding a solution $(\epsilon,S)$ of~\eqref{eqn:min-max-phi-relax},
we can round the entries of $\epsilon$ to $\{0,1\}$, while respecting the
constraint $\|\epsilon\|_{1,\infty}\leq k$
(for example, rounding the largest $k$ entries of each row to~$1$ and 
the rest entries to~$0$).
Then we can recover the prototypes using the formula~\eqref{eqn:optimal-W}.
\begin{table*}
 \caption{Comparison of test accuracy of the different classification algorithms on low dimensional ($n < 20$) OpenML datasets.  `-' indicates that GP failed to yield a solution in all the runs. $K$, in SMaLL, was set to $n$ for these data. 
 \label{tab:compare}}   
 \addtolength{\tabcolsep}{-3.5pt}
\centering
\begin{center}
 \begin{tabular} {*{12}{c}} 
  & LSVM & RF &  AB &  LR &  DT &  kNN &  RSVM & GB &  QDA &  GP &  SMaLL \\
 \hline
 \hline
 
\hline
{\bf bankruptcy} & .84$\pm$.07 & .83$\pm$.08 & .82$\pm$.05 & .90$\pm$.05 & .80$\pm$.05 & .78$\pm$.07 & .89$\pm$.06 & .81$\pm$.05 & .78$\pm$.15 & .90$\pm$.05 & {\bf .92$\pm$.06}\\
\hline
 
{\bf vineyard} & .79$\pm$.10 & .72$\pm$.06 & .68$\pm$.04 & .82$\pm$.08 & .69$\pm$.13 & .70$\pm$.11 & .82$\pm$.07 & .68$\pm$.09 & .75$\pm$.07 & .71$\pm$.12 & {\bf .83$\pm$.07}\\
\hline

{\bf pwLinear} & .83$\pm$.02 & .85$\pm$.03 & .83$\pm$.04 & .85$\pm$.02 & .80$\pm$.05 & .78$\pm$.04 & .85$\pm$.03 & .85$\pm$.03 & {\bf .87$\pm$.01} & - & .85$\pm$.02\\
\hline

{\bf sleuth1714} & .82$\pm$.03 & .82$\pm$.04 & .81$\pm$.14 & {\bf .83$\pm$.04} & {\bf .83$\pm$.06} & .82$\pm$.04 & .76$\pm$.03 & .82$\pm$.06 & .63$\pm$.13 & .80$\pm$.03 & {\bf .83$\pm$.05}\\
\hline
{\bf sleuth1605} & .66$\pm$.09 & .70$\pm$.07 & .64$\pm$.08 & .70$\pm$.07 & .63$\pm$.09 & .66$\pm$.05 & .65$\pm$.09 & .65$\pm$.09 & .62$\pm$.05 & {\bf .72$\pm$.07} & {\bf .72$\pm$.05}\\
\hline
{\bf sleuth1201} & {\bf .94$\pm$.05} & {\bf .94$\pm$.03} & .92$\pm$.05 & .93$\pm$.03 & .91$\pm$.05 & .90$\pm$.04 & .89$\pm$.09 & .88$\pm$.06 & .89$\pm$.07 & .91$\pm$.08 & {\bf .94$\pm$.05}\\
\hline
{\bf rabe266} & .93$\pm$.04 & .90$\pm$.03 & .91$\pm$.04 & .92$\pm$.04 & .91$\pm$.03 & .92$\pm$.03 & .93$\pm$.04 & .90$\pm$.04 & .94$\pm$.03 & {\bf .95$\pm$.04} & .94$\pm$.02\\
\hline
{\bf rabe148} & .95$\pm$.04 & .93$\pm$.04 & .91$\pm$.08 & .95$\pm$.04 & .89$\pm$.07 & .92$\pm$.05 & .91$\pm$.06 & .91$\pm$.08 & .92$\pm$.09 & .95$\pm$.02 & {\bf .96$\pm$.04}\\
\hline
{\bf vis\_env} & .66$\pm$.04 & .68$\pm$.05 & .66$\pm$.03 & .65$\pm$.08 & .62$\pm$.04 & .57$\pm$.03 & {\bf .69$\pm$.06} & .64$\pm$.03 & .62$\pm$.07 & .65$\pm$.09 & {\bf .69$\pm$.03}\\
\hline

{\bf hutsof99} & .74$\pm$.07 & .66$\pm$.04 & .64$\pm$.09 & .73$\pm$.07 & .60$\pm$.10 & .66$\pm$.11 & .66$\pm$.14 & .67$\pm$.05 & .59$\pm$.07 & .70$\pm$.05 & {\bf .75$\pm$.04}\\
\hline
{\bf human\_dev} & .88$\pm$.03 & .85$\pm$.04 & .85$\pm$.03 & .89$\pm$.04 & .85$\pm$.03 & .87$\pm$.03 & .88$\pm$.03 & .86$\pm$.03 & .88$\pm$.03 & .88$\pm$.02 & {\bf .89$\pm$.04}\\
\hline
{\bf fri\_c0\_100\_10} & .77$\pm$.04 & .74$\pm$.03 & .76$\pm$.03 & .77$\pm$.03 & .64$\pm$.07 & .71$\pm$.05 & {\bf .79$\pm$.03} & .71$\pm$.05 & .74$\pm$.03 & .78$\pm$.01 & .77$\pm$.06\\
\hline
{\bf elusage} & .90$\pm$.05 & .84$\pm$.06 & .84$\pm$.06 & .89$\pm$.04 & .84$\pm$.06 & .87$\pm$.05 & .89$\pm$.04 & .84$\pm$.06 & .90$\pm$.04 & .89$\pm$.04 & {\bf .92$\pm$.04}\\
\hline
{\bf diggle\_table} & .65$\pm$.14 & .61$\pm$.07 & .57$\pm$.08 & .65$\pm$.11 & .60$\pm$.09 & .58$\pm$.07 & .57$\pm$.13 & .57$\pm$.06 & .62$\pm$.07 & .60$\pm$.13 & {\bf .68$\pm$.07}\\
\hline
{\bf baskball} & .70$\pm$.02 & .68$\pm$.04 & .68$\pm$.02 & .71$\pm$.03 & .71$\pm$.03 & .63$\pm$.02 & .66$\pm$.05 & .69$\pm$.04 & .69$\pm$.04 & .68$\pm$.02 & {\bf .72$\pm$.06}\\
\hline

{\bf michiganacc} & .72$\pm$.06 & .67$\pm$.06 & .71$\pm$.05 & .71$\pm$.04 & .67$\pm$.06 & .66$\pm$.07 & .71$\pm$.05 & .69$\pm$.04 & .72$\pm$.04 & .71$\pm$.05 & {\bf .73$\pm$.05}\\
\hline
{\bf election2000} & .92$\pm$.04 & .90$\pm$.04 & .91$\pm$.03 & .92$\pm$.02 & .91$\pm$.03 & .92$\pm$.01 & .90$\pm$.07 & .92$\pm$.02 & .72$\pm$.06 & .92$\pm$.03 & {\bf .94$\pm$.02}\\
\hline
 \end{tabular}
 \end{center}
 \label{default}
 \end{table*}

\subsection{The Mirror-Prox Algorithm}

\begin{algorithm}[h]
\caption{Mirror-Prox algorithm for solving~\eqref{eqn:min-max-phi-relax}}
\label{MP-batch}
\begin{algorithmic}[1]
\State Initialize $\epsilon^{(0)}$ and $\epsilon^{(0)}$
\For{$t = 0, 1,  \ldots, T$}
\State \underline{Gradient step:}
\vspace{0.5ex}
\State $\hat{\epsilon}^{(t)} = \mathrm{Proj}_{\mathcal{E}}\bigl(\epsilon^{(t)}-\alpha_t \nabla_{\epsilon} \phi(\epsilon^{(t)}, S^{(t)})\bigr)$
\State $\hat{s}_i^{(t)}\!= \mathrm{Proj}_{\mathcal{S}_i}\bigl(
s_i^{(t)}\!+\beta_t \nabla_{s_i}\phi(\epsilon^{(t)}, S^{(t)})\bigr),~i\in[m]$
\vspace{0.5ex}
\State {\underline{Extra-gradient step:}} 
\vspace{0.5ex}
\State $\epsilon^{(t+1)} = \mathrm{Proj}_{\mathcal{E}}\bigl(\epsilon^{(t)}-\alpha_t \nabla_{\epsilon} \phi(\hat{\epsilon}^{(t)}, \hat{S}^{(t)})\bigr)$
\State $s_i^{(t+1)}\!= \mathrm{Proj}_{\mathcal{S}_i}\!\bigl(
s_i^{(t)}\!+\beta_t \nabla_{s_i}  \phi(\hat{\epsilon}^{(t)}, \hat{S}^{(t)})\bigr),~i\!\in\![m]$
\EndFor
\State $\hat{\epsilon}=\sum_{t=1}^T\alpha_t\,\hat{\epsilon}^{(t)}\big/\sum_{t=1}^T\alpha_t$
\State $\hat{S}=\sum_{t=1}^T\beta_t\,\hat{S}^{(t)}\big/\sum_{t=1}^T\beta_t$
\vspace{0.5ex}
\State Round $\hat{\epsilon}$ to $\{0,1\}^{p\times n}$
\State $\hat{W} = - \frac{1}{m\lambda} \sum_{i\in[m]} y_i(\hat{s}_i x_i^T)\odot\hat{\epsilon}$
\end{algorithmic}
\end{algorithm}  

Algorithm~\ref{MP-batch} is a customized Mirror-Prox algorithm for solving 
the convex-concave saddle-point problem~\eqref{eqn:min-max-phi-relax}, which
enjoys a $O(1/t)$ convergence rate
\citep{Nemirovski2004mirror-prox,JuditskyNemirovski2011chapter6}.

The partial gradients of $\phi(\epsilon,S)$ are given as
\begin{align*}
\nabla_{\epsilon} \phi(\epsilon,S)
&= - \frac{1}{m\lambda} 
\biggl(\sum_{i\in[m]}y_i\bigl(s_i x_i^T\bigr)\!\biggr)
\odot
\biggl(\sum_{i\in[m]}y_i\bigl(s_i x_i^T\bigr)\!\biggr), \\
\nabla_{s_i} \phi(\epsilon,S)
&= - \frac{1}{m\lambda} 
y_i \biggl(\sum_{i\in[m]}y_i\bigl(s_i x_i^T\bigr)\odot\epsilon\biggr) x_i, 
~~i\in[m].
\end{align*}
There are two projection operators in Algorithm~\ref{MP-batch}.
The first one is to project some $\epsilon\in\reals^{p\times n}$ onto
the convex set
\[
  \mathcal{E}\triangleq\left\{ \epsilon\in\reals^{p\times n} :
  \epsilon\in[0,1]^{p\times n}, ~\|\epsilon\|_{1,\infty}\leq k\right\},
\]
which can be done efficiently by Algorithm~\ref{ProjectQ}.
Essentially, we perform $p$ independent projections, 
each for one row of~$\epsilon$ using a bi-section type of algorithm
\citep{Brucker84,PardalosKovoor90,DuchiSSC08}.
We prove the following result in the appendix.

\begin{proposition} \label{ProjectingEps}
Algorithm \ref{ProjectQ} computes, up to a specified tolerance $tol$, 
the projection of any $\epsilon \in \reals^{p\times n}$ onto~$\mathcal{E}$ in $\mathcal{O}\left(\log_2(1/tol)   \right)$ time.
\end{proposition}

There are two case for the projection of $s_i\in\reals^{p}$ onto the set $S_i$.
For $i\in I_+$, we only need to project $s_{i,j(i)}$ onto the interval $[-1,0]$
and set $s_{i,j}=0$ for all $j\neq j(i)$.
For $i\in I_-$, the projection algorithm is similar to Algorithm~\ref{ProjectQ},
and we omit the details here.

For the step sizes $\alpha_t$ and $\beta_t$, they can be set according to
the guidelines described in 
\citep{Nemirovski2004mirror-prox,JuditskyNemirovski2011chapter6}, which 
depends on the smoothness properties of the function $\phi(\epsilon,S)$. 
In practice, we follow the adaptive tuning procedure developed in
\cite{JalaliFazelXiao2017VGF}.

\begin{algorithm}[h]
  \caption{$(\mathrm{Proj}_{\mathcal{E}})$ Projection of a vector onto\\ 
    \mbox{\hspace{1cm}} $\mathcal{E}_j\triangleq\{\epsilon_j\in\reals^n: \epsilon_{ji}\in[0,1], ~\|\epsilon_j\|_1\leq k\}$.\\
Given input $\epsilon_j\in\reals^n$ and a small tolerance $tol$. 
\label{ProjectQ}}
\begin{algorithmic}[1]
\State $\mbox{Clip}~ \epsilon_j~ \mbox{to} [0,1]^n: ~\hat{\eps}_{j,i} = \begin{cases}  0 & \mbox{ if } \eps_{j,i} \leq 0\\ 1 & \mbox{ if } \eps_{j,i} \geq 1\\ \eps_{j,i} & \mbox{otherwise} \end{cases}, ~i\in[n]\ $
\State Return $\hat{\eps}_j$ if ${\bf 1}^T\hat{\eps}_j \leq k$
\Statex \hspace*{0.5cm} \underline{Do binary search until $tol$-solution found}
\State Set $low = \left({\bf 1}^T \eps_j - k\right)/n$
\State Set $high = \max_{i \in [n]} \eps_{j,i} - k/n$
\While{$low \leq high$}   
\State Set $\lambda = (low + high)/2$    
\State Compute $\hat{\epsilon_j}: \forall i \in [n], \hat{\eps}_{j,i} = \eps_{j,i} - \lambda$
\State Clip $\hat{\epsilon}_j$ to $[0, 1]^n$
\If{$|{\bf 1}^T \hat{\epsilon}_j - k| < tol$}  
\State return $\hat{\eps}_j$ 
\ElsIf{${\bf 1}^T \hat{\epsilon}_j > k$}
\State Set $low = (low + high)/2$  
\Else 
\State Set $high = (low + high)/2$ 
\EndIf
\EndWhile
\end{algorithmic}
\end{algorithm}


\begin{figure*}
\centering
\includegraphics[width=0.9\textwidth, trim = {0cm 0cm 0cm 0cm}, clip]{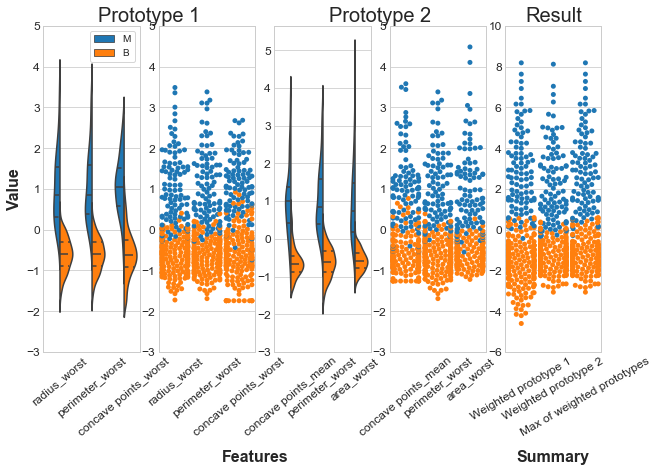} 
\caption{SMaLL applied to the Breast Cancer dataset with $k=3$ and $p=2$. The blue and the orange dots represent the test instances from the two classes. The plots show the kernel density estimates and the actual values of the non-zero features in each prototype, as well at the final predictor result.}\label{fig:violin}
\end{figure*}

\section{Experiments}\label{sec:experiments}
We demonstrate the merits of our approach with a set of experiments. First, we attempt to give some intuition on how the class of sparse multiprototype linear predictors differs from other popular model classes. Figure \ref{fig1:sleuth} is a visualization of the decision surface of different types of classifiers on the 2-dimensional \emph{chscase funds} toy dataset, obtained from OpenML. The two classes are shown in red and blue, with training data in solid shade and test data in translucent shade. The color of each band indicates the gradation in the confidence of prediction - each classifier is more confident in the darker regions and less confident in the lighter regions. The $2$-prototype linear predictor attains the best test accuracy on this toy problem ($0.73$). Note that some of the examples are highlighted by a black rectangle - the linear classifiers (logistic regression and linear SVM) could not distinguish between these examples, whereas the $2$-prototype linear predictor was able to distinguish and assign them to different bands.

\subsection{Low-dimensional Datasets Without Sparsity}

Next, we compare the accuracy of SMaLL with $k = n$ (no sparsity) to the accuracy of other standard classification algorithms, on various low-dimensional ($n \leq 20$) binary classification datasets from OpenML. The methods that we compare against are: linear SVM (LSVM), SVM with non-linear kernels such as radial basis function, polynomial, and sigmoid  (RSVM), Logistic Regression (LR), Decision Trees (DT), Random Forest (RF), $k$-Nearest Neighbor (kNN), Gaussian Process (GP), Gradient Boosting (GB), AdaBoost (AB), and Quadratic Discriminant Analysis (QDA). All the datasets were normalized so that each feature has zero mean and unit variance. Since the datasets do not specify separate train, validation, and test sets, we measure test accuracy by averaging over five random train-test splits. We pre-clustered the positive examples into $p = 2$ clusters, and initialized the prototypes with the cluster centers. 

We trained parameters by $5$-fold cross-validation. The coefficient of the error term $C$ in LSVM and $\ell_2$-regularized LR was selected from $\{0.1, 1, 10, 100\}$. In the case of RSVM, we also added $0.01$ to the search set for $C$, and chose the best kernel between a radial basis function (RBF), polynomials of degree 2 and 3, and sigmoid. For the ensemble methods (RF, AB, GB), the number of base predictors was selected from the set $\{10, 20, 50\}$. The maximum number of features for RF estimators was optimized over the square root and the log selection criteria. We also found best validation parameters for DT (gini or entropy for attribute selection), kNN (1, 3, 5 or 7 neighbors), and GP (RBF kernel scaled with scaled by a coefficient in the set $\{0.1, 1.0, 5\}$ and dot product kernel with inhomogeneity parameter $\sigma$ set to 1). Finally, for our algorithm SMaLL, we fixed $\lambda = 0.1$ and $\alpha_{t} = 0.01$, and we searched over $\beta_t = \beta \in \{0.01, 0.001\}$. \\ \\

Table \ref{tab:compare} shows the test accuracy for the different algorithms on different datasets. As seen from the table, SMaLL with $k=n$ generally performed very well on most of these datasets. This substantiates the practicality of SMaLL in the low dimensional regime.  

\subsection{Higher-Dimensional Datasets with Sparsity}
On higher-dimensional data, feature selection becomes critical. To substantiate our claim that our technique produces an interpretable model, we ran SMaLL on the Breast Cancer dataset with $k=3$ and $p=2$ (two prototypes, three non-zero elements in each). Figure \ref{fig:violin} shows the kernel density estimate plots and the actual values of the non-zero features in each prototype, as well at the final predictor result. Note that the feature {\em perimeter\_worst} appears in both prototypes. As the rightmost plot shows, the predictor output provides a good separation of the test data, and SMaLL registered a test accuracy of over $94\%$. It is straightforward to understand how the resulting classifier reaches its decisions: which features it relies on and how those features interact. 

\begin{figure*}
\centering
\includegraphics[height=9.5cm, trim = {2cm 0cm 0cm 0cm}, clip]{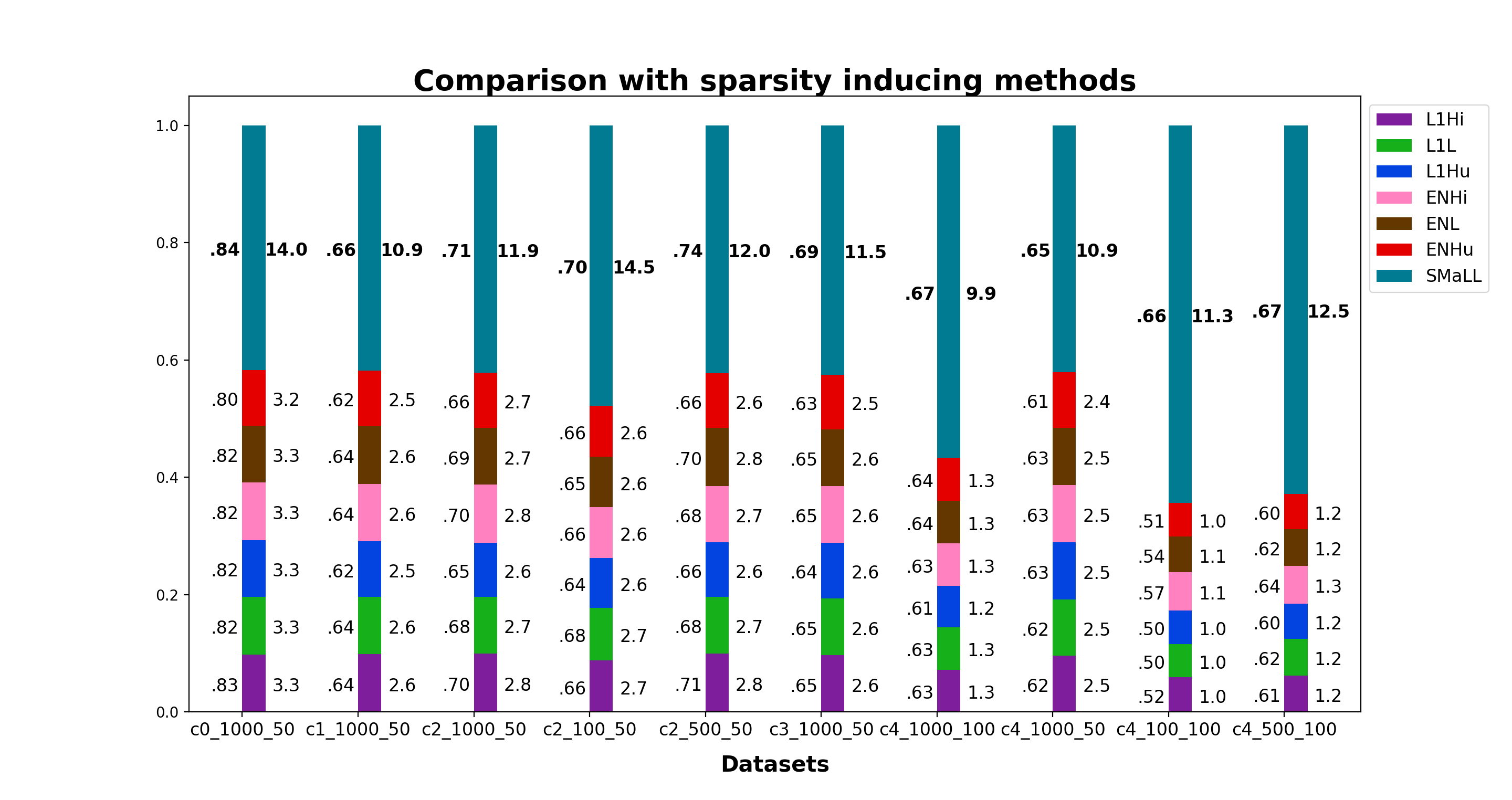} 
\caption{{\bf Comparison on high dimensional ($n >= 50$) OpenML datasets.} Each stacked bar shows two numbers: average test accuracy on the left, and the ratio of average test accuracy in \% to the  number of selected features on the right. \label{fig:sparse}}       
\end{figure*}



Next, we compare SMaLL with six other methods, which induce sparsity by minimizing an $\ell_1$-regularized loss function. These methods minimize one of three empirical loss functions (hinge loss, log loss, and the binary-classification Huber loss), regularized by either an $\ell_1$ or an elastic net penalty (i.e. $\ell_1$  {\it and} $\ell_2$). We refer to these as L1Hi ($\ell_1$, hinge),  L1L ($\ell_1$, log), L1Hu ($\ell_1$, Huber), EnHi (elastic net, hinge), ENL (elastic net, log) and ENHu (elastic net, huber). 

Note that while we can explicitly control the amount of sparsity in SMaLL, the methods that use $\ell_1$ or Elastic Net regularization do not have this flexibility. Therefore, in order to get the different baselines on the same footing, we devised the following empirical methodology. We trained each baseline and selected the $n/2$ features with the largest absolute values. Then, we retrained the classifier using only the selected features, using the same loss (hinge, loss, or log) and an $\ell_2$ regularization. Our procedure ensured that each baseline benefited, in effect, from an {\em elastic net}-like regularization while having the most important features at its disposal. For the SMaLL classifier, we fixed $k = 3$ and $p = 2$. As before, when the original dataset did not specify a train-test split, our results are averaged over five random splits.   The parameters for each method were tuned using a 5-fold grid search cross-validation procedure. We fixed $\lambda = 0.1$ and $\eta = 1e-4$, and performed a joint search over $\alpha_{t} \in \{0.1, 1e-2, 1e-3\}$ and $\beta_t \in \{1e-3, 1e-4\}$. For all the baselines, we optimized the cross validation error over the $\ell_1$ regularization coefficients in the set  $\{1e-1, 1e-2, 1e-3, 1e-4\}$. Moreover, in case of elastic net, the ratio of the $\ell_1$ coefficient to the $\ell_2$ coefficient was set to 1.

Figure \ref{fig:sparse} provides empirical evidence that SMaLL yields extremely sparse yet more accurate prototypes than the baselines on several high dimensional OpenML datasets. The first number in each dataset name indicates the number of examples, and the second the dimensionality.  In SMaLL, since some features might be selected in more than one prototype, we included the multiplicity while computing the total feature count.  Each bar in the plot is scaled to unit length, with each method getting a share proportional to its normalized accuracy on the dataset. The normalized accuracy indicates the effective information conveyed per selected feature.  We see that SMaLL outperforms other methods in both accuracy as well as the normalized performance, the latter by an order of magnitude.  This shows the promise of convex relaxations of SMaLL toward achieving succinct yet accurate weight representation in the high dimensional regime.  In fact, as Fig. \ref{fig:sparse} shows, the savings in terms of sparsity could be truly  remarkable.  This observation is further reinforced in Fig. \ref{fig:fri}. We observe that SMaLL registers a normalized value over 14, which is significantly higher than all the other methods, without incurring a significant dip in test accuracy compared to the most accurate method (GB).     

\clearpage
\bibliographystyle{authordate1}
\bibliography{bibfile}
\clearpage


\appendix
\clearpage
\section{Supplementary Material}
{\bf Proof of Proposition \ref{prop:conjugate}}
\begin{proof}
By definition, the Fenchel conjugate
$$u^*(s) = \sup_{t\in\reals^p} \left(\sum_{k=1}^p s_k t_k - \log\left(1 + \sum_{k=1}^{p} \exp(-t_k)\right)\right).$$  
Equating the partial derivative with respect to each $t_k$ to 0, we get 
\begin{equation} \label{s}
s_k ~=~ - \dfrac{\exp(-t_k^*)}{1 + \sum_{c = 1}^{p}  \exp(-t_c^\ast)}~~,
\end{equation}
or equivalently,
\begin{equation*}\label{t}
t_k^* = - \log \left(-s_k \left(1 + \sum_{c = 1}^{p} \exp(-t_c^*)\right) \right) ~~.
\end{equation*}
We note from \eqref{s} that
$$\dfrac{1}{1 + \sum_{c=1}^p \exp(-t_c^*)} =  1 + s^{\top} {\bf 1}~~.$$ 
Using the convention $0 \log 0 = 0$, the form of the conjugate function
in~\eqref{eqn:u-conjugate} can be obtained by plugging $t^{*} = (t_1^*, \ldots, t_r^*)$ into $u^*(s)$ and performing some simple algebraic manipulations.  

Proposition~\ref{prop:conjugate} follows directly from the form of $u^*$,
especially the constraint set $\mathcal{S}_i$ for $i\in I_-$.
For $i\in I_+$, we notice that the conjugate of $\ell(z)=\log(1+\exp(-z))$
is
\[
  \ell^*(\beta) = (-\beta)\log(-\beta) + (1+\beta)\log(1+\beta), \quad
  \beta\in[-1,0].
\]
Then we can let the $j(i)$th entry of $s_i\in\reals^p$  be $\beta\in[-1,0]$
and all other entries be zero.
Then we can express $\ell^*$ through $u^*$ as shown in the proposition.
\end{proof}

\vspace{2ex}

{\bf Proof of Proposition \ref{prop:eliminate-W}}
\begin{proof}

Recall that 
\[
\Phi(W, \epsilon, S) \!=\! \frac{1}{m} \!\! 
\sum_{i\in[m]} \!\Bigl(y_i s_i^T(W\odot\epsilon)x_i - u^*(s_i)\!\Bigr)
 + \dfrac{\lambda}{2} ||W||_F^2 .
\]
Then 
$$\nabla_W \Phi(W,\epsilon,S) \!=\! \frac{1}{m} \!\! 
\sum_{i\in[m]} y_i \bigl(s_i x_i^T\bigr)\odot\epsilon + \lambda W~~.$$
The proof is complete by setting $\nabla_W \Phi(W,\epsilon,S)=0$, 
and solving for~$W$.
\end{proof}
\vspace{5ex}

{\bf Proof of Proposition \ref{ProjectingEps}}
\begin{proof}
In order to project $a\in\reals^n$ onto 
\[
\mathcal{E}_j\triangleq\{\epsilon_j\in\reals^n: \epsilon_{ji}\in[0,1], ~\|\epsilon_j\|_1\leq k\},
\]
we need to solve the following problem:
\begin{eqnarray*}
\min_{x \in \reals^{n}} & \dfrac{1}{2} ||x - a||^2 \\
\text{s.t.} \qquad & \sum_{i=1}^d x_i \leq k\\
\forall i \in [n]: & 0 \leq x_i \leq 1~~.
\end{eqnarray*}
Our approach is to form a Lagrangian and then invoke the KKT conditions. Introducing Lagrangian parameters $\lambda \in \reals_{+}$ and $u, v \in \reals_{+}^{d}$,  we get the Lagrangian $L(x, \lambda, u, v)$
\begin{eqnarray*}
& = & \dfrac{1}{2} ||x - a||^2 + \lambda \left(\sum_{i=1}^n x_i - k\right) -  \sum_{i=1}^n u_i x_i  \\
 & \qquad \qquad + & \sum_{i=1}^n v_i(x_i - 1)\\
& = & \dfrac{1}{2} ||x - a||^2 + \sum_{i=1}^n x_i(\lambda - u_i + v_i) \\
& \qquad \qquad - & \lambda k - \sum_{i=1}^n v_i~~.
\end{eqnarray*}
Therefore,
\begin{equation} \label{GradLag}
\nabla_{x^*} L = 0 \implies x^* = a - (\lambda \mathbf{1} - u + v)~~.
\end{equation}

We note that $g(\lambda, u, v) \triangleq L(x^*, \lambda, u, v)$
\begin{eqnarray*}
 =~  - \dfrac{1}{2} ||\lambda \mathbf{1} - u + v||^2 + a^{\top} (\lambda \mathbf{1} - u + v) - \lambda K - \mathbf{1}^{\top} v. \label{g_function}
\end{eqnarray*}

Using the notation $b \succeq t$ to mean that each coordinate of vector $b$ is at least $t$,  our dual is
\begin{equation} \label{LagDual}
\max_{\lambda \geq 0, u \succeq 0, v \succeq 0}  g(\lambda, u, v)~~.
\end{equation}

We now list all the KKT conditions:
\begin{eqnarray*}
\forall i \in [n]: \qquad x_i > 0 & \implies &  u_i = 0\\
\forall i \in [n]: \qquad x_i < 1 & \implies &  v_i = 0\\
\forall i \in [n]: \qquad u_i > 0 & \implies &  x_i = 0\\
\forall i \in [n]: \qquad v_i > 0 & \implies &  x_i = 1~~~~~~~.\\
\forall i \in [n]: \qquad u_i v_i & = & 0\\
\sum_{i=1}^n x_i < k & \implies & \lambda = 0\\
\lambda > 0 & \implies & \sum_{i=1}^n x_i = k\\
\end{eqnarray*}

\begin{figure*}
\centering
\includegraphics[height=8.2cm, trim = {2cm 0cm 0cm 0cm}, clip]{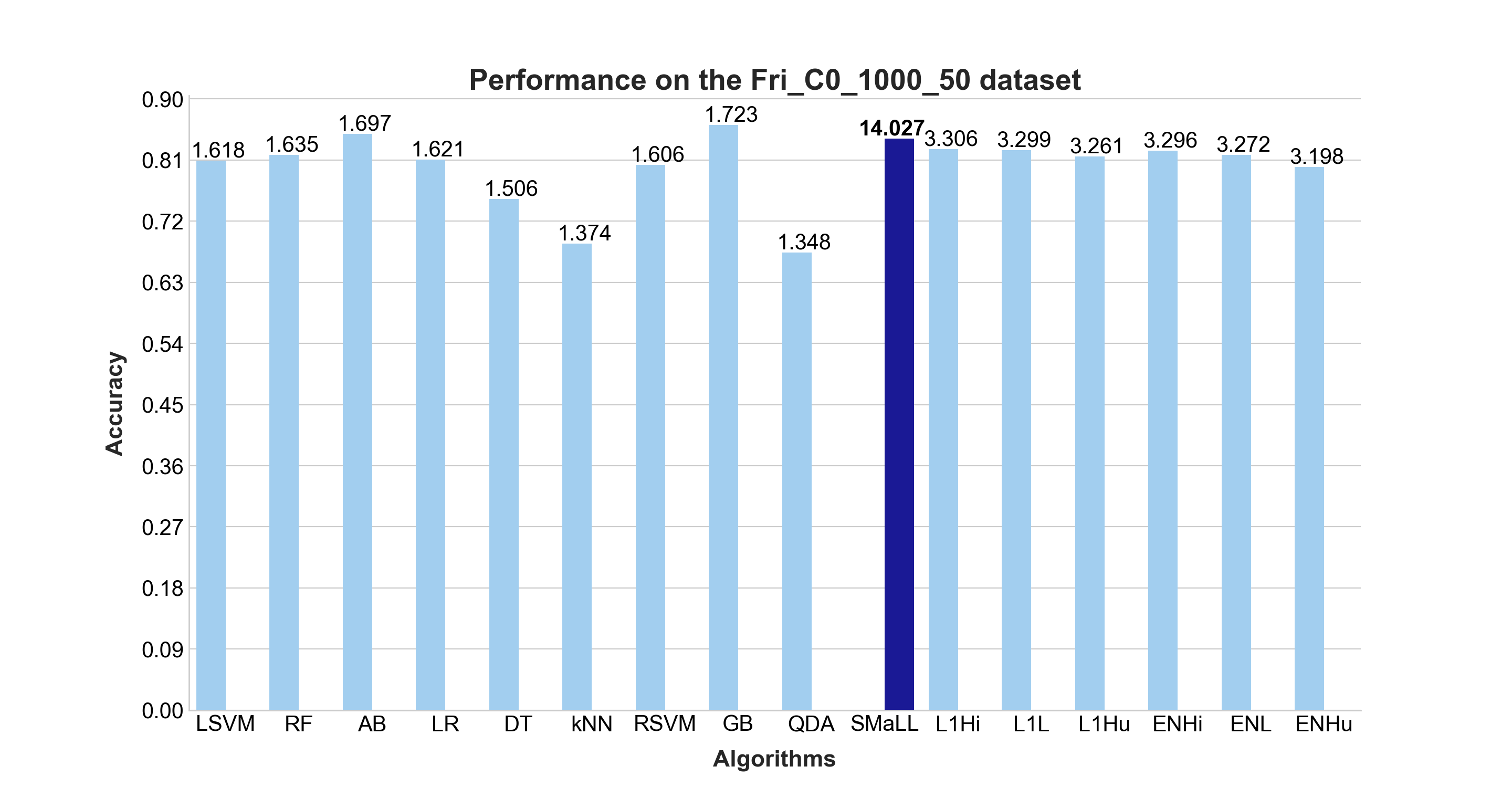} 
\caption{{\bf The big picture.} The plot depicts the performance of SMaLL to both the standard classification algorithms and the sparsity inducing baselines on the fri\_c0\_1000\_50 dataset. The dataset consists of 1000 examples in 50 dimensions. The number atop each bar is the ratio of average test accuracy in percentage to the average number of selected features. 
\label{fig:fri}}
\end{figure*}

We consider the two cases, (a) $\sum_{i=1}^n x^*_i < k$, and (b) $\sum_{i=1}^n x^*_i = k$ separately. \\

First consider $\sum_{i=1}^n x^*_i < k$. Then, by KKT conditions, we have the corresponding $\lambda = 0$. 
Consider all the sub-cases. Using \eqref{GradLag}, we get
\begin{enumerate}
\item $x^*_i = 0 \implies a_i = \lambda - u_i + v_i = - u_i \leq 0$ (since $x^*_i < 1$, therefore, by KKT conditions, $v_i = 0$). 
\item $x^*_i = 1 \implies a_i = 1 + \lambda - u_i + v_i = 1 + v_i \geq 1$ (since $x^*_i > 0$, therefore, $u_i = 0$ by KKT conditions).   
\item $0 < x^*_i < 1 \implies a_i = x^*_i + \lambda - u_i + v_i = x^*_i$. 
\end{enumerate} 
Now consider $\sum_{i=1}^n x^*_i = k$. Then, we have $\lambda \geq 0$. Again, we look at the various sub-cases. 
\begin{enumerate}
\item $x^*_i = 0  \implies a_i = \lambda - u_i + v_i = \lambda - u_i \implies u_i = -(a_i - \lambda)$. Here, $u_i$ denotes the amount of clipping done when $a_i$ is negative.   
\item $x^*_i = 1  \implies a_i = 1 + \lambda - u_i + v_i =  1 + \lambda + v_i \implies  v_i = -(1 + \lambda - a_i)$.  Here, $v_i$ denotes the amount of clipping done when $a_i > 1$.  Also, note that $a_i \geq 1$ in this case. 
\item $0 < x^*_i < 1 \implies a_i = x^*_i + \lambda - u_i + v_i = x^*_i + \lambda \implies x^*_i = a_i - \lambda$. In order to determine the value of $\lambda$, we note that since $\sum_{i=1}^n x^*_i = k$, therefore, 
\begin{eqnarray*} \sum_{i=1}^n (a_i - \lambda) = k \implies \sum_{i=1}^n a_i - n \lambda = k \\
\implies \lambda = \dfrac{1}{n} \sum_{i=1}^n a_i - \dfrac{k}{n}  \leq \max_i a_i - \dfrac{k}{n}~~. 
\end{eqnarray*}
\end{enumerate}

Algorithm \ref{ProjectQ} implements these different sub-cases and thus accomplishes the desired projection. The algorithm is a bisection method, and thus converges linearly to a solution within the specified tolerance $tol$. 
\end{proof}




\end{document}